\documentclass[twoside,leqno,twocolumn]{article}

\usepackage[letterpaper]{geometry}

\usepackage{array}
\usepackage{ltexpprt}
\usepackage{graphicx}
\usepackage{multirow}
\usepackage{bm}
\usepackage{caption}
\usepackage{subfigure}
\usepackage{algorithmic}
\usepackage{amsmath,amssymb,amsfonts}
\usepackage{textcomp}
\usepackage{xcolor}
\usepackage{makecell}
\usepackage{url}
\newcommand{\stitle}[1]{\vspace{1mm}\noindent\textbf{#1\ \ }}

\begin{document}

\newcommand\relatedversion{}
\renewcommand\relatedversion{\thanks{The full version of the paper can be accessed at \protect\url{https://arxiv.org/abs/1902.09310}}} 

\title{Abnormal Event Detection via Hypergraph Contrastive Learning}

\author{Bo Yan \thanks{Beijing University of Posts and Telecommunications, China. Email: \{boyan, yangcheng, shichuan, liu\_jiawei, 2018211801\}@bupt.edu.cn} \and Cheng Yang \footnotemark[1] \and Chuan Shi \footnotemark[1] \footnote{Corresponding author} \and Jiawei Liu \footnotemark[1] \and Xiaochen Wang \footnotemark[1]}

\date{}

\maketitle


\fancyfoot[R]{\scriptsize{Copyright \textcopyright\ 2023 by SIAM\\
Unauthorized reproduction of this article is prohibited}}





\begin{abstract} \small\baselineskip=9pt 
Abnormal event detection, which refers to mining unusual interactions among involved entities, plays an important role in many real applications. Previous works mostly over-simplify this task as detecting abnormal pair-wise interactions. However, real-world events may contain multi-typed attributed entities and complex interactions among them, which forms an Attributed Heterogeneous Information Network (AHIN). With the boom of social networks, abnormal event detection in AHIN has become an important, but seldom explored task. In this paper, we firstly study the unsupervised abnormal event detection problem in AHIN. The events are considered as star-schema instances of AHIN and are further modeled by hypergraphs. A novel hypergraph contrastive learning method, named AEHCL, is proposed to fully capture abnormal event patterns. AEHCL designs the intra-event and inter-event contrastive modules to exploit self-supervised AHIN information. The intra-event contrastive module captures the pair-wise and multivariate interaction anomalies within an event, and the inter-event module captures the contextual anomalies among events. These two modules collaboratively boost the performance of each other and improve the detection results. During the testing phase, a contrastive learning-based abnormal event score function is further proposed to measure the abnormality degree of events. Extensive experiments on three datasets in different scenarios demonstrate the effectiveness of AEHCL, and the results improve state-of-the-art baselines up to 12.0\% in Average Precision (AP) and 4.6\% in Area Under Curve (AUC) respectively.
\end{abstract}

\section{Introduction}

\begin{figure}[t]
    \centering
    \includegraphics[scale=.38]{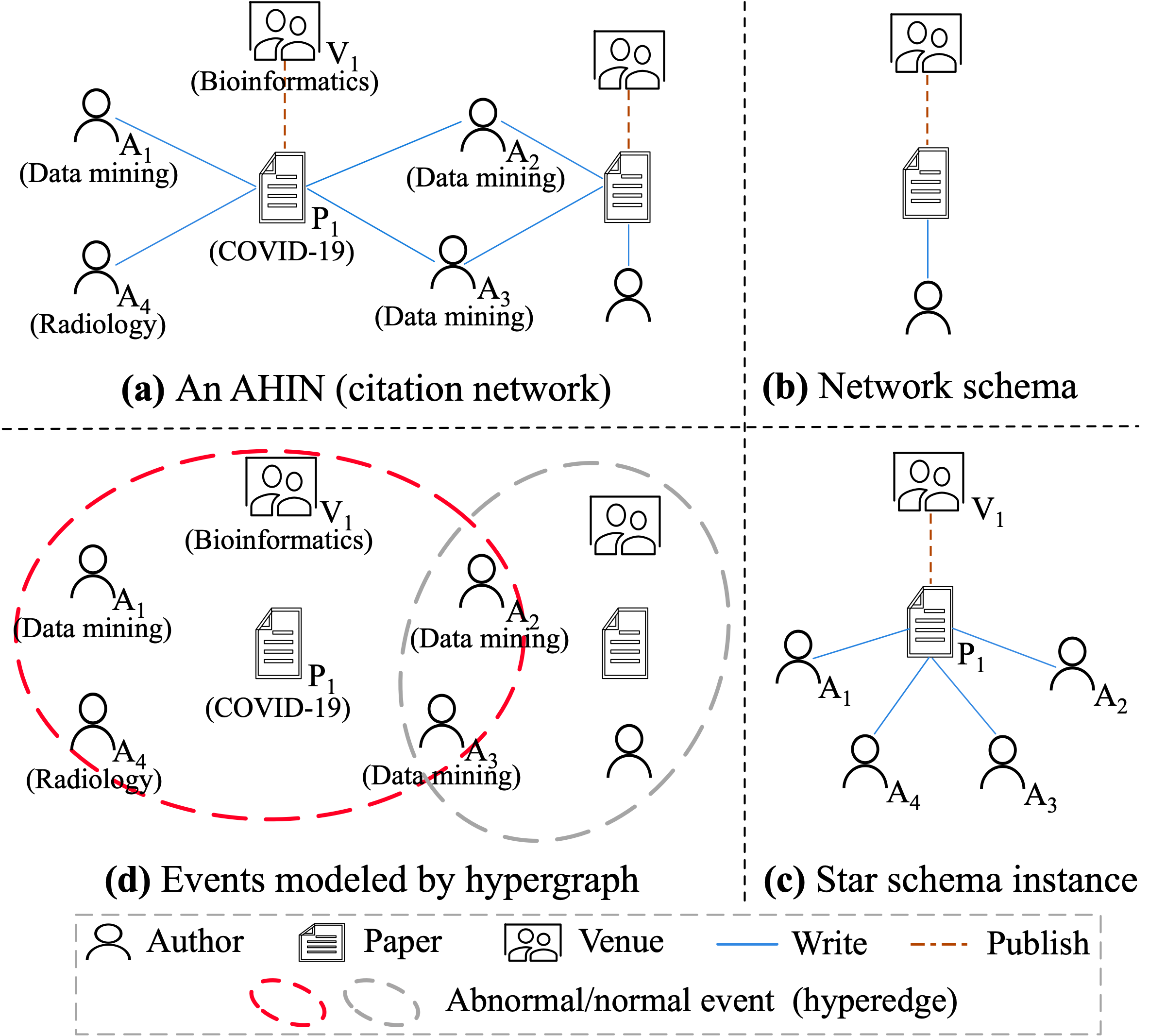}    \caption{A toy example of abnormal events in AHIN. (a) A citation network. (b) Network schema of the citation network. (c) A star schema instance (event) guided by network schema. (d) Events modeled by hypergraph.}
    \label{ab_example}
\vspace{-6mm}
\end{figure}

Events widely exist in social networks, which consist of several entities and complex interactions among them. Abnormal events refer to unusually or rarely happened events among observed event samples. Figure \ref{ab_example}(d) shows a toy example of abnormal events in the citation network, which is considered abnormal since the involved entities exhibit rare interactions (data-mining specialists seldom collaborate on a COVID-19 paper with radiologists). To date, abnormal event detection has become a significant task in fraud detection \cite{DBLP:conf/ijcnn/ZhengZWPSG18}, Internet of things (IoT) security \cite{DBLP:conf/kdd/WangCNLCT21}, computer network monitoring \cite{DBLP:conf/ijcai/ChenTSCZ16}, and other real applications, thus drawing increasing concerns in recent years.

Existing approaches mainly focus on detecting events with pair-wise abnormal patterns \cite{DBLP:conf/cikm/FanSW18, DBLP:conf/ijcai/ChenTSCZ16, DBLP:conf/cikm/AkogluTVF12}. Moreover, some studied events only have simple structures \cite{DBLP:conf/icdm/WangZWDZW18, DBLP:conf/ijcnn/ZhengZWPSG18} or categorical attributes \cite{DBLP:conf/kdd/WangCNLCT21, DBLP:conf/bigdataconf/HanCXY21}. However, real-world events may contain multiple types of entities with rich attributes, as well as complex interactions among them, which forms an Attributed Heterogeneous Information Network (AHIN) \cite{DBLP:conf/aaai/HuZSZLQ19}. As depicted in Figure \ref{ab_example}(a), the event of publishing a paper in AHIN is associated with a set of multi-typed attributed nodes, making the event not limited to simple pair-wise interactions (e.g., a data mining specialist writes a COVID-19 paper), but complex set-wise interactions (e.g., several data mining specialists cooperate to write a COVID-19 paper on a bioinformatics venue). Detecting abnormal events in AHIN became an essential, but largely neglected task in social network analysis and abnormal detection. For example, detecting abnormal collaborations in the academic network can better understand research trends and promote cooperation.


Although this task is highly urged to be explored, it's never a trivial task. First, the complex set-wise patterns in AHIN make the modeling of abnormal events more challenging. Traditional pair-wise modeling loses its effects since all pair-wise interactions may be normal but the set-wise interaction is abnormal. In this regard, a general framework should be proposed to model events in AHIN, as well as fully consider complex abnormal set-wise patterns. Second, due to the scarcity of anomalies and the prohibitive cost of labeling processes, we need to conduct abnormal event detection in an unsupervised manner, which compounds the difficulties of capturing abnormal patterns in AHIN. Finally, to measure the abnormal degrees in an unsupervised setting, a proper abnormal event score function is desired, which is capable of truly reflecting the abnormal degrees. 

To tackle these challenges, in this paper, we study the problems of unsupervised abnormal event detection in AHIN and propose a novel framework for \underline{\textbf{A}}bnormal \underline{\textbf{E}}vent detection via \underline{\textbf{H}}ypergraph \underline{\textbf{C}}ontrastive \underline{\textbf{L}}earning (AEHCL). First, we consider an event in AHIN as a star-schema instance \cite{DBLP:journals/tkde/ShiLZSY17} and further use the hyperedge \cite{DBLP:journals/tkde/GuiLTJNK017, DBLP:conf/aaai/TuCWW018} to model the set-wise interactions. A hyperedge in the hypergraph is associated with more than two entities, endowing the entities within an event not limited to pair-wise interactions. Second, based on this event modeling, we propose a hypergraph contrastive learning method to fully capture abnormal patterns in an unsupervised manner, which includes two contrastive strategies at the intra-event and inter-event levels. At the intra-event level, the pair-wise contrastive module captures abnormal interactions between node pairs within an event, and the multivariate contrastive module further considers high-order interactions. At the inter-event level, we contrast events with their neighbors to capture the inconsistent patterns between local events. All these modules are optimized in an end-to-end manner. Finally, to measure the abnormal degrees of events in AHIN, we further propose a novel contrastive-based abnormal event score function, which effectively integrates the detection results of the above-mentioned modules.

The contributions are summarized as follows:

(1) To the best of our knowledge, this is the first attempt to study the problem of abnormal event detection in AHIN, which is an important and practical task in many application scenarios. 

(2) We propose a novel hypergraph contrastive learning method, called AEHCL, leveraging both intra- and inter-event contrastive tasks to capture abnormal event patterns within an event and between events. A contrastive-based abnormal event score function is further designed to measure the abnormal degrees.

(3) We conduct extensive experiments on three types of AHIN datasets. The results improve state-of-the-art baselines up to 12.0\% in AP and 4.6\% in AUC scores respectively, which demonstrate the effectiveness of our model.

\section{Related Work}

\subsection{Abnormal Event Detection}
The abnormal event usually implies that a set of entities as well as their interactions rarely co-occurrence. For a set of entities with abnormal structures, \cite{DBLP:conf/icdm/WangZWDZW18} detects suspicious dense blocks by utilizing an anomaly-aware representation method. \cite{DBLP:conf/ijcnn/ZhengZWPSG18} further maps different types of nodes into a shared latent space. These models haven't formally defined abnormal events and also ignore rich attributes of entities. Some other works study discrete event sequence abnormal detection [3, 27, 10], where each event is represented by a categorical value \cite{DBLP:conf/kdd/WangCNLCT21, DBLP:conf/bigdataconf/HanCXY21}. The event in the HIN is first defined by \cite{DBLP:journals/tkde/GuiLTJNK017} as a node set that forms a complete semantic unit. \cite{DBLP:conf/ijcai/ChenTSCZ16} proposes APE to model the weighted pair-wise interactions of the heterogeneous categorical event \cite{DBLP:conf/cikm/AkogluTVF12}. AEHE \cite{DBLP:conf/cikm/FanSW18} further incorporates node attributes and node's second-order representations to detect abnormal events with specific meta-path. Besides these abnormal event detection studies, the most relevant task is known as abnormal node detection. CoLA \cite{DBLP:journals/tnn/LiuLPGZK22}, and ANEMONE \cite{DBLP:conf/cikm/JinLZCLP21} are two advanced works using contrastive learning to model mismatching patterns between neighbor nodes. Both these works only model abnormal nodes or abnormal events with simple patterns but fail to capture complex abnormal interactions of events in AHIN.

\subsection{Heterogeneous Information Network Analysis}
Heterogeneous information networks (HIN) \cite{DBLP:journals/tkde/ShiLZSY17} analysis focuses on modeling complex entities and their rich relations in various applications. They mainly leverage meta-path to learn semantic-preserving representations of HIN. Metapath2vec \cite{DBLP:conf/kdd/DongCS17} designs a meta-path based random walk and utilizes skip-gram to learn HIN embedding. HAN \cite{DBLP:conf/www/WangJSWYCY19} proposes node and semantic-level attention to learn the importance of meta-paths. HeCo \cite{DBLP:conf/kdd/WangLHS21} introduces a cross-view contrastive mechanism to HIN analysis. HeteHG-VAE \cite{DBLP:journals/pami/FanZWLZGD22} further uses hypergraphs to model high-order complex interactions of HIN and proposes a hypergraph variational auto-encoder to learn robust node representations. HIN analysis has been widely used in fraud detection \cite{DBLP:conf/aaai/HuZSZLQ19}, recommendation \cite{DBLP:conf/kdd/Lu0S20}, and other fields. However, few works use HIN analysis in abnormal event detection.

\subsection{Contrastive Learning}
Contrastive learning has achieved great success in representation learning \cite{DBLP:conf/icml/ChenK0H20, DBLP:conf/cvpr/He0WXG20}, the core idea of which is to contrast positive pairs against negative pairs. As for graph-related contrastive learning methods, DGI \cite{DBLP:conf/iclr/VelickovicFHLBH19} proposes an unsupervised learning objective based on mutual information theory to contrast, MVGRL \cite{DBLP:conf/icml/HassaniA20} maximizes the mutual information between the node representation of one view and the graph representation of another view. In the heterogeneous graph domain, DMGI~\cite{DBLP:conf/aaai/ParkK0Y20} introduces a consistency regularisation framework that minimized divergence between the specific relation type of node embeddings. Besides, some works apply contrastive learning to abnormal node detection \cite{DBLP:journals/tnn/LiuLPGZK22, DBLP:journals/corr/abs-2108-09896}. However, there is no research on using contrastive learning for abnormal event detection in AHIN.

\section{Preliminary}
In this paper, we perform abnormal event detection in the Attributed Heterogeneous Information Network (AHIN), which can be formally defined as follows:


\stitle{Definition 1. Attributed Heterogeneous Information Network (AHIN).} An AHIN is denoted as $G=(V, E, X)$ consisting of an object set $V$, a link set $E$ and an attribute matrix $X \in R^{|V|\times k}$. It is also associated with a object type mapping function $\phi$ $:$ $V \rightarrow \mathcal{A}$ and a link type mapping function $\varphi$ $:$ $E \rightarrow \mathcal{R}$. $\mathcal{A}$ and $\mathcal{R}$ denote the sets of predefined object and link types, where $|\mathcal{A}|+|\mathcal{R}|>2$. The $\textbf{network schema}$ \cite{DBLP:journals/tkde/ShiLZSY17} $S^G=(\mathcal{A}, \mathcal{R})$ can be seen as a meta template of an AHIN $G$, which is a graph defined over object types $\mathcal{A}$, with links as relations from $\mathcal{R}$.

\stitle{Definition 2. Event in AHIN.} An event $Q=(v^q, V^q, E^q, X^q, S^G)$ in an AHIN is an instance of star schema $S^G$, which includes an object set $V^q \subseteq V$ , a link set $E^q \subseteq E$ and an attribute matrix $X^q \in R^{|V^q| \times k}$.  $v^q \in V^q$ is the \textbf{center node} which uniquely identifies an event $Q$ and the \textbf{context node} set $V^c=V^q \backslash v^q$ denotes the context of an event.

A toy example of AHIN and its network schema is shown in Figure \ref{ab_example}(a) and (b) respectively for the citation network. Particularly, star schema network is a kind of AHIN, in which links only exist between objects with a center type and others. Guided by star schema, we can extract star schema instances (events) from the AHIN. Figure \ref{ab_example}(c) shows an example of paper publication event. The center node is a paper uniquely identifying the event, and the context nodes are the venue and authors. Instead of modeling each event as a set of binary relations (links), we employ hypergraph \cite{DBLP:books/daglib/0067501} to model events by viewing involved nodes as a whole.

\stitle{Definition 3. Hypergraph Modeling for Event.} A hypergraph $G^h=(V^h, E^h)$ consists of nodes $V^h=V$ from the AHIN $G$ and a $\textbf{hyperedge}$ set $E^h$. Each hyperedge $E^h_i \in E^h$ connects nodes in $V^q$ of an event $Q$. The center node $v_q$ uniquely identifies a hyperedge.

Thus, an event can be modeled as a hyperedge in the hypergraph, which is capable of modeling the high-order semantics of an event (e.g., multiple authors are co-authored in one paper). An event is abnormal if the event exhibits rare interaction patterns. As depicted in Figure \ref{ab_example}(d), the publication events are modeled by hyperedges, which involve multiple types of nodes (i.e., paper, author, and venue), and the abnormal event contains the semantics that data mining specialists cooperate with a radiologist on a COVID-19 paper, which is rarely happened and considered as an anomaly.

\stitle{Definition 4. Abnormal Event Detection in AHIN.} Given an AHIN $G=(V, E, X)$ and an event set $\mathcal{Q}=\{Q_1,\ldots,Q_m\}$, the goal is to learn an anomaly score function $f_{\theta}(\cdot)$ to obtain the anomaly score $s_i=f_{\theta}(Q_i)$ for $\forall Q_i \in \mathcal{Q}$. By choosing the optimal threshold $t$ of anomaly score, the abnormal event set $\mathcal{Q}^{\prime}=\{Q_i|Q_i \in \mathcal{Q}, f_\theta(Q_i) \geq t\}$ can be detected.

\section{The Proposed Model}
\begin{figure*}[!t]
    \centering
    \includegraphics[scale=.37]{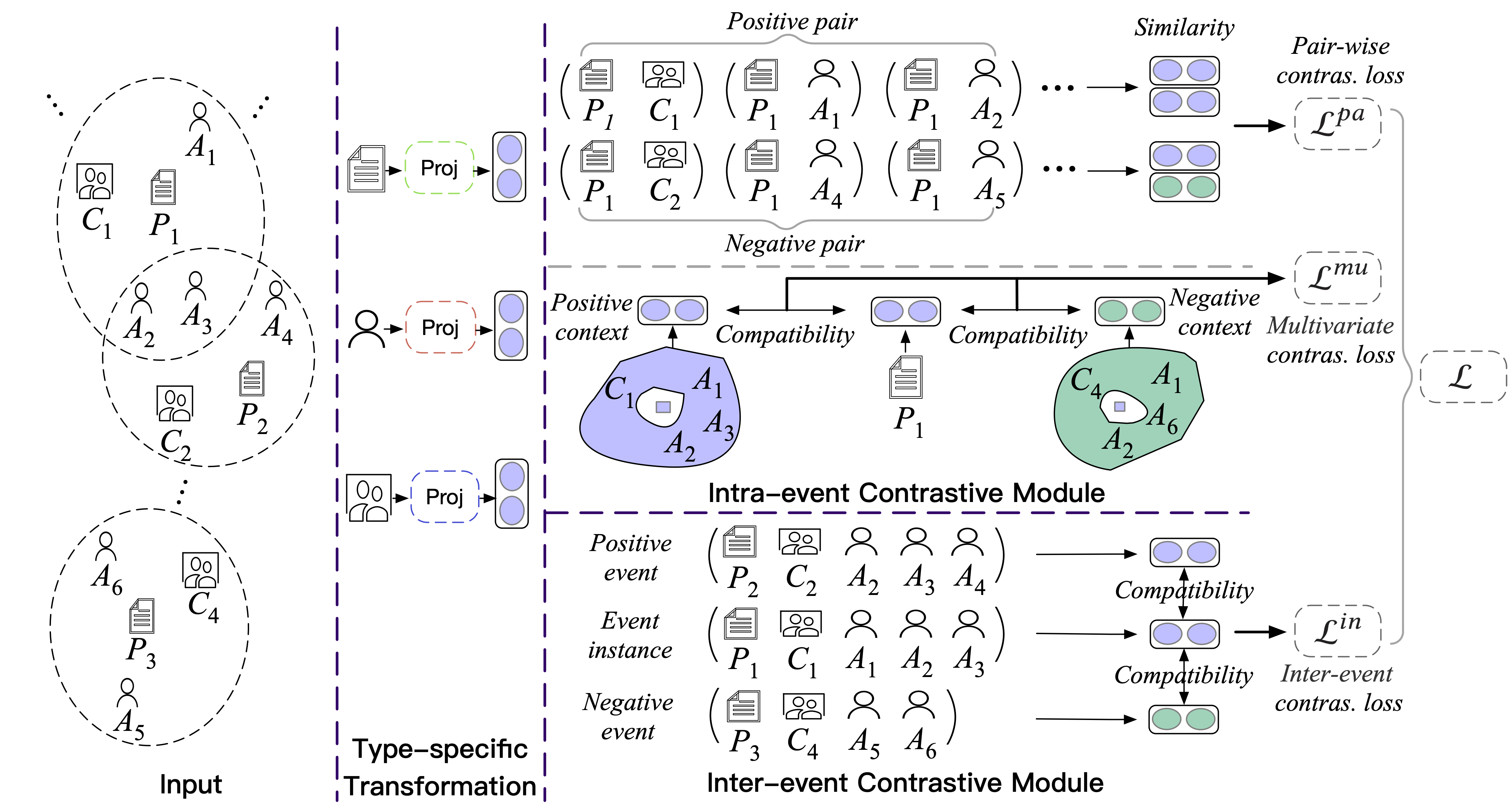}
    \caption{The overall architecture of AEHCL.}
    \label{model_overall}
\vspace{-6mm}
\end{figure*}

In this section, we present AEHCL, a novel hypergraph contrastive learning method for abnormal event detection. The overall architecture is depicted in Figure \ref{model_overall}. To fully capture abnormal event patterns in AHIN, we design contrastive modules at the intra-event and inter-event levels. For intra-event, we model the node-pair similarity within an event to capture abnormal interactions of two nodes. High-order abnormal interactions are further captured by the multivariate contrastive task between the center and context nodes. For inter-event, we define the neighbors of events and design contrastive tasks among neighbor events. Based on these contrastive tasks, a novel abnormal event score function is proposed to measure the overall abnormal degrees. 

\subsection{Type-specific Node Transformation}
Since an event contains multiple types of nodes, directly using the original features for downstream tasks will suffer from reduced performance \cite{DBLP:conf/emnlp/HuYSJL19}. A solution is performing heterogeneous graph convolution \cite{DBLP:conf/emnlp/HuYSJL19}. However, the aggregation operation may harm the original feature interaction patterns. Therefore, for a node $v_i$ with type $\phi_i$, we directly transform its original feature $x_i$ into a shared latent space via a type-specific transformation matrix $W_{\phi_i}$, i.e., $z_i = \sigma(W_{\phi_i} \cdot x_i+b_{\phi_i})$, where $b_{\phi_i} \in \mathbb{R}^{d}$ is a vector bias. $\sigma(\cdot)$ denotes activation function. After transformation, the representation of each node lies in a shared space with $d$ dimension.

\subsection{Intra-event Contrastive Module}
The intra-event contrastive module aims to capture abnormal patterns within an event. These patterns are presented by complicated interactions between involved nodes, which can be classified into two types: pair-wise and multivariate interaction patterns. Accordingly, we design pair-wise and multivariate contrastive modules to capture these patterns.

\subsubsection{Pair-wise Contrastive Module} Pair-wise proximity has been used in hypergraph representation learning methods \cite{DBLP:conf/aaai/TuCWW018}. The basic intuition behind them is that the pair-wise node matching degree within a hyperedge (event) should be higher than others, which inspires us to model the normal matching patterns of node pairs. Then the node pairs that do not conform to this pattern are considered anomalies. Existing methods directly fuse all pair-wise matching degrees within an event to obtain an event abnormal score. However, this operation may weaken the true abnormal degrees. Therefore, we separately model each node's matching patterns with others rather than fusing them. Specifically, for a node $v_j$ in an event $Q_i$, we optimize the following contrastive loss:
\vspace{-1mm}
\begin{equation}
\mathcal{L}_j^{pa} = -\log \frac{\sum_{v_k \in \mathcal{P}^{pa}_j} \exp(sim(z_j,z_k)/\tau)}{\sum_{v_l \in \{\mathcal{P}^{pa}_j \cup \mathcal{N}^{pa}_j\}}\exp(sim(z_j, z_l)/\tau)},
	\label{equ:pcm}
\vspace{-1mm}
\end{equation}
where $sim(\cdot)$ is a matching function which is the cosine similarity in our model and $\tau$ is the temperature parameter. $\mathcal{P}^{pa}_j = \{v_k|v_k \in V_i^q \backslash v_j\}$ is the positive node set of node $v_j$ and we randomly sampled $n$ nodes do not belong to $Q_i$ as the negative node set $\mathcal{N}^{pa}_j$. Finally, we sum all node's contrastive losses within an event and average losses among all events to get the final pair-wise contrastive loss:
\vspace{-1mm}
\begin{equation}
    \mathcal{L}^{pa} = \frac{1}{m}\sum_{i=1}^{m}\sum_{v_j \in V^q_i} \mathcal{L}^{pa}_j.
	\label{equ:pcm_total}
\vspace{-1mm}
\end{equation}

By minimizing $\mathcal{L}^{pa}$, the representations of node pairs within an event are similar, i.e., node pairs within a normal event has higher matching degrees compared with abnormal events.

\subsubsection{Multivariate Contrastive Module} 
The pair-wise contrastive module only captures simple abnormal pair-wise interactions, such as the rare cooperation of two authors. However, there may also exist many complicated multivariate abnormal patterns in the reality. In these cases, the pair-wise interactions within an event are all normal but the event is abnormal when considering interactions between more than two nodes. We model this kind of abnormal event by a multivariate contrastive module. The module captures the multivariate interaction patterns within an event by modeling the compatibility between the center node and context nodes. This compatibility of normal events is higher than abnormal events. For example, a paper's content is highly correlated with the type of published venue and the authors' interests. Specifically, for each node $v_i$ with type $\phi_i$, we first add a type embedding $o_{\phi_i}$ to obtain the type-aware representation $h_i$, i.e., $ h_i = z_i + o_{\phi_i}$. Type embedding makes the model aware of the interactions with heterogeneity to capture more meaningful patterns. 

To model the multivariate interactions among context node set $V^c_i$ of event $Q_i$ and obtain the final context representation $c_i^{mu}$, we apply self-attention \cite{DBLP:conf/nips/VaswaniSPUJGKP17} followed by a max-pooling operation: 
\begin{equation}
    c_i^{mu} = {\rm max \text{-} pooling}(\textsc{Att}(\{h_j|v_j \in V^c_i\})),
    \label{equ:context_emb}
\end{equation}
where the max-pooling operation is performed over each dimension of all context nodes' representations. Therefore, $c_i^{mu}$ preserves the most important interaction patterns of context nodes for identifying abnormal events. 

After that, we use a simple bilinear scoring function to model the compatibility between the center node and context nodes:
\vspace{-1mm}
\begin{equation}
    s_i^{mu} = \sigma(h_i^{q \top} W_{mu} c_i^{mu}),
    \label{equ:context_s}
\vspace{-1mm}
\end{equation}
where $h_i^{q}$ is the center node representation of the event $Q_i$ and $\sigma(\cdot)$ is the sigmoid activation function. Here, the score $s_i^{mu}$ of normal events should be close to 1, while abnormal events close to 0. Since there is no prior knowledge about abnormal events, we construct the negative context to mimic incompatibility. Instead of randomly replacing context nodes, we first use K-means algorithm to perform node clusters based on original features, then for each type of context nodes, we randomly select one node and replace it with another node with the same type but different clusters. In this way, the constructed negative context is different but not easily distinguishable from the positive context, which enhances the capacity for abnormal detection. Then the negative context representation $\tilde{c}_i^{mu}$ and corresponding bilinear score $\tilde{s}_i^m$ can be obtained similarly using Equation (\ref{equ:context_emb}) and Equation (\ref{equ:context_s}).  Finally, we adopt a standard binary cross-entropy (BCE) loss function:

\vspace{-5mm}
\begin{equation}
    \mathcal{L}^{mu} = -\frac{1}{m}\sum_{i=1}^{m}\log(s_i^{mu})+\log(1-\tilde{s}_i^{mu}).
    \label{equ:mcm_total}
\vspace{-2mm}
\end{equation}

\subsection{Inter-event Contrastive Module}
Abnormal event patterns may not only limit to abnormal node interactions within an event but also appear among events. Like incompatibility anomaly between local neighbor nodes \cite{DBLP:journals/tnn/LiuLPGZK22}, we believe that there also exists incompatibility anomaly between local neighbor events. Intuitively, the normal events are prone to have similar semantic meanings with their neighbor events, while the abnormal events are not. Based on this assumption, we design contrastive tasks to model the compatibility between neighbor events. Specifically, we first use an attention layer to obtain the event representation. Given an event $Q_i$, a type-specific attention parameter $P_{\phi_j}\in \mathbb{R}^{d \times d}$ is applied to each context node $v_j \in V^c_i$ with type $\phi_j$ to obtain attention keys:
\vspace{-2mm}
\begin{equation}
    k_j = P_{\phi_j} \cdot z_j,
    \label{equ:attention_k}
\vspace{-2mm}
\end{equation}
then the attention weights of context nodes are calculated through the softmax function:
\vspace{-2mm}
\begin{equation}
    \alpha_j = \frac{\exp(\sigma(k_j^{\top} \cdot z_i^{q}))}{\sum_{l=1}^{|V^c_i|}\exp(\sigma(k_l^{\top} \cdot z_i^{q}))},
    \label{equ:attention_w}
\vspace{-2mm}
\end{equation}
where $z_i^{q}$ is the representation of the center node. The context representation can be obtained by a weighted sum of all context node embeddings with the learned attention weights:
\vspace{-2mm}
\begin{equation}
    c_i^{in} = \sum_{j=1}^{|V^c_i|} \alpha_j z_j,
    \label{equ:attention_aggre}
\vspace{-2mm}
\end{equation}
then we concatenate the context representation $c^{in}_i$ and center node representation $z_i^{q}$ to obtain the event representation $e_i$:
\vspace{-2mm}
\begin{equation}
    e_i = c_i^{in} \| z_i^{q}.
    \label{equ:event_rep}
\vspace{-2mm}
\end{equation}
Next, we define the neighbor event set $\mathcal{P}^{in}_i$, i.e., positive sample set of event $Q_i$. Motivated by \cite{DBLP:conf/kdd/WangLHS21}, if two events are connected by more meta-paths \cite{DBLP:journals/tkde/ShiLZSY17}, they may have more similar semantic meanings. Formally, the $\mathcal{P}^{in}_i$ is defined as follows:
\vspace{-2mm}
\begin{equation}
    \mathcal{P}^{in}_i = \{Q_j|N_{i,j}^{mp}>t^{pos}\},
    \label{equ:P_a}
\vspace{-2mm}
\end{equation}
where $N_{i,j}^{mp}$ denotes the number of meta-paths between event $Q_i$ and $Q_j$. That is, when the number of meta-paths between two events exceeds threshold $t^{pos}$, then these two events are positive samples of each other. Similarly, we define the negative sample set $\mathcal{N}^{in}_i$ when the number of shared nodes is less than threshold $t^{neg}$. With the positive set $\mathcal{P}^{in}_i$ and negative set $\mathcal{N}^{in}_i$, we define the bilinear scoring function to model the compatibility between neighbor events:

\vspace{-5mm}
\begin{equation}
    s_i^{in} = \sigma(e_i^{\top} W_{in} e_i^p),
    \label{equ:event_s}
\vspace{-2mm}
\end{equation}
where $e_i^p$ is the representation of $Q_i$'s positive sample which is randomly sampled from $\mathcal{P}^{in}_i$. The score $\tilde{s}_i^{in}$ of the negative event pair can be obtained in the same way. Then, we have the following inter-event contrastive loss:

\vspace{-5mm}
\begin{equation}
    \mathcal{L}^{in} = -\frac{1}{m}\sum_{i=1}^{m}\log(s_i^{in})+\log(1-\tilde{s}_i^{in}).
    \label{equ:intercm_total}
\vspace{-2mm}
\end{equation}

\subsection{Optimization and Prediction}
In the training stage, we optimize the above-mentioned three modules jointly. The overall optimization function is:

\vspace{-5mm}
\begin{equation}
    \mathcal{L} = \alpha*\mathcal{L}^{pa}+\beta*\mathcal{L}^{mu}+\gamma*\mathcal{L}^{in},
    \label{equ:objfunc}
\vspace{-2mm}
\end{equation}
where $\alpha$, $\beta$, and $\gamma$ are the trade-off parameters to balance three modules. We will discuss them in detail in experiments. The training flow of AEHCL can be found in the supplementary material.

After the model is well trained, we use the following abnormal score function to measure the abnormal degree of an event $Q_i$:
\vspace{1mm}
\begin{equation}
    f_{\theta}(Q_i)=-(\alpha*\min \limits_{v_j,v_k \in V_i^q}(sim(z_j,z_k))+\beta*s_i^{mu}+\gamma*s_i^{in}),
    \label{equ:score}
\vspace{-2mm}
\end{equation}
where $\alpha$, $\beta$, and $\gamma$ are the same as used in Eq. (\ref{equ:objfunc}). $\min (\cdot)$ operation means that we use the minimal node-pair similarity within an event to measure the pair-wise interaction abnormal degree, since the event is likely abnormal only if one node-pair similarity is relatively small. Another choice is the sum of all node-pair similarity scores, which is infeasible since events may have different numbers of entities. We also test other operations (e.g., average) and detailed results are shown in experiments. $s_i^{mu}$ and $s_i^{in}$ mean that we use bilinear scores of positive pairs to measure the multivariate and inter-event abnormal degrees. The bilinear scores of abnormal events are relatively small since they have poor local compatibility. Finally, we add these three scores to obtain the final abnormal event score $s_i=f_{\theta}(Q_i)$. Since we take negative signs, a large abnormal score $s_i$ indicates the event is more likely abnormal.

\section{Experiments\protect\footnote{More details can be found in the supplementary material}}
\subsection{Experimental Setup}\

\stitle{Datasets.} To evaluate the performance of the proposed method, we employ three real-world AHIN datasets. The academic network dataset Aminer, the movie network dataset IMDB, and the takeaway orders dataset Meituan. The basic information of them is summarized in Table~\ref{tab:data}.




\begin{table}[h]
\vspace{-2mm}
\renewcommand{\arraystretch}{1.8} 
  \centering
  \fontsize{12}{12}\selectfont
  \caption{Dataset statistics.}
  \resizebox{\columnwidth}{!}{
    \begin{tabular}{c|c|c|c|c|c}
    \hline
    \textbf{Dataset} & \textbf{\makecell*[c]{Center \\nodes}} & \textbf{\makecell*[c]{Context \\Nodes}} & \textbf{\makecell*[c]{\# Avg. \\nodes}} & \textbf{\# Events} & \textbf{\makecell*[c]{\# Abnormal \\events}} \\
    \hline
    \hline
    Aminer & \makecell*[l]{Paper: 20567} & \makecell*[l]{Author: 13541\\Venue: 115 }& 4.3 & 20567 & 1028 \\ \hline
    IMDB & \makecell*[l]{Movie: 4661} & \makecell*[l]{Director: 2240\\ Actor: 5749} & 5 & 4661 & 233 \\ \hline
    Meituan & \makecell*[l]{Order: 24330} & \makecell*[l]{Shop: 2265\\ User: 6972} & 3 & 24330 & 1216 \\ \hline
    \end{tabular}%
    }
\label{tab:data}%
\vspace{-4mm}
\end{table}%

\begin{table*}[ht]
    \renewcommand\arraystretch{1.2}
    \centering
    \caption{Overall performance (\%$\pm \sigma$) of AEHCL on abnormal event detection on three datasets. The best result of all models is in bold and the best result of previous models is underlined. The improvement (\%) is based on the underlined results.}
\resizebox{\textwidth}{3.5cm}{
    \begin{tabular}{c|cc|cc|cc}
\hline
                 \textbf{Datasets}& \multicolumn{2}{c|}{\textbf{Aminer}}                                         & \multicolumn{2}{c|}{\textbf{IMDB}}                                           & \multicolumn{2}{c}{\textbf{Meituan}}                                         \\ \hline
                 \hline
\textbf{Metrics} & \multicolumn{1}{c|}{AP}           & AUC           & \multicolumn{1}{c|}{AP}            & AUC          & \multicolumn{1}{c|}{AP}            & AUC           \\ \hline

Metapath2vec     & \multicolumn{1}{c|}{19.3$\pm$1.2 \enspace $-$60.9}         & 73.1$\pm$1.3 \enspace $-$13.7          & \multicolumn{1}{c|}{16.2$\pm$0.9 \enspace $-$74.3}           & 57.9$\pm$1.9 \enspace $-$37.2          & \multicolumn{1}{c|}{39.1$\pm$0.0 \enspace $-$37.0}           & 85.5$\pm$1.0 \enspace $-$4.3\hphantom{0}          \\ \hline

HeCo             & \multicolumn{1}{c|}{21.9$\pm$0.0 \enspace $-$55.6}          & 79.8$\pm$0.0 \enspace $-$5.8\hphantom{0}      & \multicolumn{1}{c|}{19.1$\pm$0.2 \enspace $-$69.7}          & 62.6$\pm$0.7 \enspace $-$32.1        & \multicolumn{1}{c|}{—}                    & —                 \\ \hline

HeteHG-VAE       & \multicolumn{1}{c|}{23.0$\pm$0.0 \enspace $-$53.3}         & 55.2$\pm$0.0 \enspace $-$34.8           & \multicolumn{1}{c|}{11.2$\pm$0.0 \enspace $-$69.7}          & 52.5$\pm$0.0 \enspace $-$35.5         & \multicolumn{1}{c|}{12.8$\pm$1.0 \enspace $-$79.4}          & 63.4$\pm$1.3 \enspace $-$29.0          \\ \hline

ANEMONE          & \multicolumn{1}{c|}{37.7$\pm$0.4 \enspace $-$23.5}          & 82.0$\pm$0.3 \enspace $-$3.2\hphantom{0}            & \multicolumn{1}{c|}{55.7$\pm$0.1 \enspace $-$11.7}          & \underline{92.2}$\pm$0.0 \enspace $-$\hphantom{20.0}         & \multicolumn{1}{c|}{55.5$\pm$0.1 \enspace $-$10.6}          & \underline{89.3}$\pm$0.0 \enspace $-$\hphantom{20.0}         \\ \hline
CoLA             & \multicolumn{1}{c|}{33.2$\pm$0.3 \enspace $-$32.7}          & 83.5$\pm$0.2 \enspace $-$1.4\hphantom{2}          & \multicolumn{1}{c|}{\underline{63.1}$\pm$0.0 \enspace $-$\hphantom{20.0}}           & 91.8$\pm$0.1 \enspace $-$0.4\hphantom{0}         & \multicolumn{1}{c|}{\underline{62.1}$\pm$0.0 \enspace $-$\hphantom{20.0}}          & 86.6$\pm$0.1 \enspace $-$3.0\hphantom{0}          \\ \hline

APE              & \multicolumn{1}{c|}{\underline{49.3}$\pm$0.0 \enspace $-$\hphantom{20.0}}          & \underline{84.7}$\pm$0.0  \enspace $-$\hphantom{20.0}        & \multicolumn{1}{c|}{30.0$\pm$0.0 \enspace $-$52.5}              & 81.2$\pm$0.0 \enspace $-$12.0          & \multicolumn{1}{c|}{30.7$\pm$0.8 \enspace $-$50.6}          & 83.0$\pm$1.1 \enspace $-$7.1\hphantom{0}          \\ \hline

AEHE             & \multicolumn{1}{c|}{48.1$\pm$0.0 \enspace $-$2.4\hphantom{0}}         & 84.6$\pm$0.0 \enspace $-$0.1\hphantom{0}          & \multicolumn{1}{c|}{28.3$\pm$0.0 \enspace $-$55.2}           & 76.9$\pm$0.0 \enspace $-$16.6         & \multicolumn{1}{c|}{40.2$\pm$0.2 \enspace $-$35.3}          & 85.6$\pm$1.1 \enspace $-$4.1\hphantom{0}           \\ \hline \hline

$\text{AEHCL}_{intra}$    & \multicolumn{1}{c|}{52.2$\pm$0.2 \enspace $+$5.9\hphantom{0}} & 88.0$\pm$0.1 \enspace $+$3.9\hphantom{0}&  \multicolumn{1}{c|}{65.3$\pm$0.0 \enspace $+$3.5\hphantom{0}}& \textbf{95.7$\pm$0.0} \enspace \textbf{$+$3.8}\hphantom{0.}& \multicolumn{1}{c|}{65.9$\pm$0.2 \enspace $+$6.1\hphantom{0}} & 88.0$\pm$0.1 \enspace $-$1.5\hphantom{0} \\ \hline

$\text{AEHCL}_p$    & \multicolumn{1}{c|}{32.0$\pm$0.2 \enspace $-$35.1} & 78.3$\pm$0.1 \enspace $-$7.6\hphantom{0}&  \multicolumn{1}{c|}{50.8$\pm$0.1 \enspace $-$19.5} & 93.8$\pm$0.1 \enspace $+$1.7\hphantom{0} & \multicolumn{1}{c|}{33.2$\pm$0.4 \enspace $-$46.5} & \textbf{91.6$\pm$0.1} \enspace \textbf{$+$2.6}\hphantom{0.} \\ \hline

$\text{AEHCL}_m$    & \multicolumn{1}{c|}{49.5$\pm$0.0 \enspace $+$0.4\hphantom{0}} & 86.0$\pm$0.0 \enspace $+$1.5\hphantom{0}&  \multicolumn{1}{c|}{38.9$\pm$0.1 \enspace $-$38.4} & 80.2$\pm$0.0 \enspace $-$13.0& \multicolumn{1}{c|}{59.3$\pm$0.1 \enspace $-$4.5\hphantom{0}} & 84.3$\pm$0.0 \enspace $-$5.6\hphantom{0} \\ \hline

$\text{AEHCL}_{p+inter}$    & \multicolumn{1}{c|}{33.2$\pm$0.1 \enspace $-$32.7}& 79.4$\pm$0.0 \enspace $-$6.3\hphantom{0}&  \multicolumn{1}{c|}{54.0$\pm$0.0 \enspace $-$14.4} & 93.6$\pm$0.0 \enspace $+$1.5\hphantom{0} & \multicolumn{1}{c|}{33.0$\pm$0.3 \enspace $-$46.9} & 79.8$\pm$0.1 \enspace $-$10.6 \\ \hline

$\text{AEHCL}_{m+inter}$    & \multicolumn{1}{c|}{50.2$\pm$0.0 \enspace $+$1.8\hphantom{0}} & 86.8$\pm$0.0 \enspace $+$2.5\hphantom{0}&  \multicolumn{1}{c|}{42.1$\pm$0.0 \enspace $-$33.3} & 85.5$\pm$0.0 \enspace $-$7.3\hphantom{0} & \multicolumn{1}{c|}{61.6$\pm$0.1 \enspace $-$0.8\hphantom{0}} & 89.0$\pm$0.0 \enspace $-$0.3\hphantom{0} \\ \hline

$\text{AEHCL}_{inter}$    & \multicolumn{1}{c|}{22.7$\pm$0.4 \enspace $-$54.0} & 77.9$\pm$0.1 \enspace $-$5.7\hphantom{0}&  \multicolumn{1}{c|}{12.7$\pm$0.2 \enspace $-$79.9} & 73.5$\pm$0.0 \enspace $-$20.3 & \multicolumn{1}{c|}{34.6$\pm$0.2 \enspace $-$44.3} & 76.5$\pm$0.2 \enspace $-$14.3 \\ \hline

AEHCL    & \multicolumn{1}{c|}{\textbf{55.2$\pm$0.1} \enspace \textbf{$+$12.0}\hphantom{.}} & \textbf{88.6$\pm$0.1} \enspace \textbf{$+$4.6}\hphantom{0.}&  \multicolumn{1}{c|}{\textbf{67.2$\pm$0.0} \enspace \textbf{$+$6.5}\hphantom{0.}} & 95.1$\pm$0.0 \enspace $+$3.1\hphantom{0} & \multicolumn{1}{c|}{\textbf{66.3$\pm$0.1} \enspace \textbf{$+$6.8}\hphantom{0.}} & 90.8$\pm$0.0 \enspace $+$1.7\hphantom{0} \\ \hline
\end{tabular}}
\vspace{-4mm}
    \label{tab:overall}
\end{table*}

\stitle{Abnormal Event Injection.}
In previous works \cite{DBLP:conf/ijcai/ChenTSCZ16, DBLP:conf/cikm/FanSW18}, the artificial abnormal events are generated as follows: for each event $Q_i$, $n$ target nodes are sampled from its context node set, where $n$ is sampled from $\{1, 2, 3\}$. These target nodes are then replaced by other sampled nodes with the same type. However, all baselines suffer catastrophic performance under this injection strategy. It may be that this simple injection method makes abnormal events less discriminative with normal ones due to the complex interactions of attributed nodes within an event. Therefore, similar to \cite{DBLP:journals/tnn/LiuLPGZK22}, for each target node, we sample other $k$ candidate nodes of the same type and calculate the Euclidean distance between each candidate node's attributes and the target node’s. Then, we select the node with the largest Euclidean distance to replace the target node, which makes the injected abnormal events more discriminative.

\stitle{Baselines.}
We compare our proposed model with three kinds of baselines: abnormal event detection models (APE \cite{DBLP:conf/ijcai/ChenTSCZ16}, AEHE \cite{DBLP:conf/cikm/FanSW18}); graph abnormal detection models (CoLA \cite{DBLP:journals/tnn/LiuLPGZK22}, ANEMONE \cite{DBLP:conf/cikm/JinLZCLP21}; graph representation learning models (metapath2vec \cite{DBLP:conf/kdd/DongCS17}, HeCo \cite{DBLP:conf/kdd/WangLHS21}, HeteHG-VAE \cite{DBLP:journals/pami/FanZWLZGD22}). We also design variants of AEHCL, including intra-event contrastive module $\text{AEHCL}_{intra}$, pair-wise contrastive module $\text{AEHCL}_{p}$, multivariate contrastive module $\text{AEHCL}_{m}$, inter-event contrastive module $\text{AEHCL}_{inter}$ and combinations of different modules:
$\text{AEHCL}_{p+inter}$ and $\text{AEHCL}_{m+inter}$.


\stitle{Implementation Details.}
For all of the baselines, we randomly run 10 times to report the average results. We tune all hyper-parameters to report the best performance. For abnormal detection models, we test all feasible abnormal event score functions and report the best. For AEHCL, we use Adam optimizer and the learning rate is set as 0.001. For optimizing the pair-wise contrastive module solely, we decay the learning rate for every two epochs. The hidden dimension is set as 64, and the original feature dimensions are 108, 128, and 300 for Aminer, IMDB, and Meituan datasets respectively. The temperature parameter $\tau$ is set to 1 for all datasets. The hyperparameters $\alpha$, $\beta$, and $\gamma$ are set as \{1, 0.8, 0.2\}, \{1, 0.1, 0.1\}, and \{0.5, 1, 0.3\} in Aminer, IMDB, and Meituan datasets respectively.

\subsection{Overall Performance}
\label{sec:overall}
Following previous works \cite{DBLP:conf/cikm/FanSW18}, we use two popular evaluation metrics AP and AUC. The overall results are reported in Table \ref{tab:overall}. Note that HeCo suffers memory issues in the Meituan dataset due to its full-batch training. We have the following observations: (1) The proposed AEHCL model and its variants outperform all the baselines on all the datasets. The improvements are up to 12.0\% in AP and 4.6\% in AUC, which demonstrates the effectiveness of our model for abnormal event detection in AHIN by fully considering different abnormal event patterns. (2) Generally, only keeping individual module of AEHCL (e.g., $\text{AEHCL}_p$ and $\text{AEHCL}_m$) yields poor results, especially in AP scores, while the combination of different modules can obtain more gains. Although adding the inter-event contrastive module will slightly harm the AUC in IMDB and Meituan dataset, it improves much AP score. It's a trade-off between AP and AUC scores and the weights between modules can be flexibly adjusted under different conditions. In brief, different modules are complementary to each other and focus on capturing different patterns of abnormal events. (3) Graph representation learning methods perform poorly on all datasets, indicating that pure graph representation learning is far from enough to capture abnormal event patterns. The contrastive learning method, HeCo, also performs worse, which shows the challenge of applying contrastive learning to abnormal event detection. The performance of existing abnormal node/event detection models significantly declines in one or two datasets, since they only focus on capturing simple patterns of anomalies (e.g., abnormal attributes of the individual node or abnormal pair-wise interactions).

\begin{table}[!t]
\vspace{2mm}
\renewcommand{\arraystretch}{1.7} 
  \centering
  \fontsize{7.5}{7.5}\selectfont
  \caption{Abnormal event cases detected by AEHCL on Aminer and Meituan datasets (dm: data mining).}
  \resizebox{\columnwidth}{1.8cm}{
    \begin{tabular}{c|m{6.4cm}<{\raggedright}}
    \hline
    \multicolumn{1}{c|}{\textbf{Dataset}} & \multicolumn{1}{c}{\textbf{Abnormal events}} \\
    \hline
    \hline
    \multirow{1}{*}[-5pt]{Aminer} &
    \makecell*[l]{ Paper:\enspace Text processing (COVID-19) \\
          Author(s): \enspace Jiawei Han (dm), Xuan Wang (dm), \\Kang Zhou (dm) \\
          Conf: \enspace BIBM (bio-medicine) }\\
    \hline
    \multirow{1}{*}[-5pt]{Meituan} & 
    \makecell*[l]{Order A: \enspace price $>$ 50, taste id=1 \\
          Shop B: \enspace category id=1 \\
          User C: \enspace price (avg.) $<$ 29, taste id (avg.)=2, \\category id=3} \\
    \hline
    \end{tabular}%
    }
\vspace{-7mm}
\label{tab:case_ab}%

\end{table}%

\begin{figure*}[ht]
    \centering
    \includegraphics[scale=0.255]{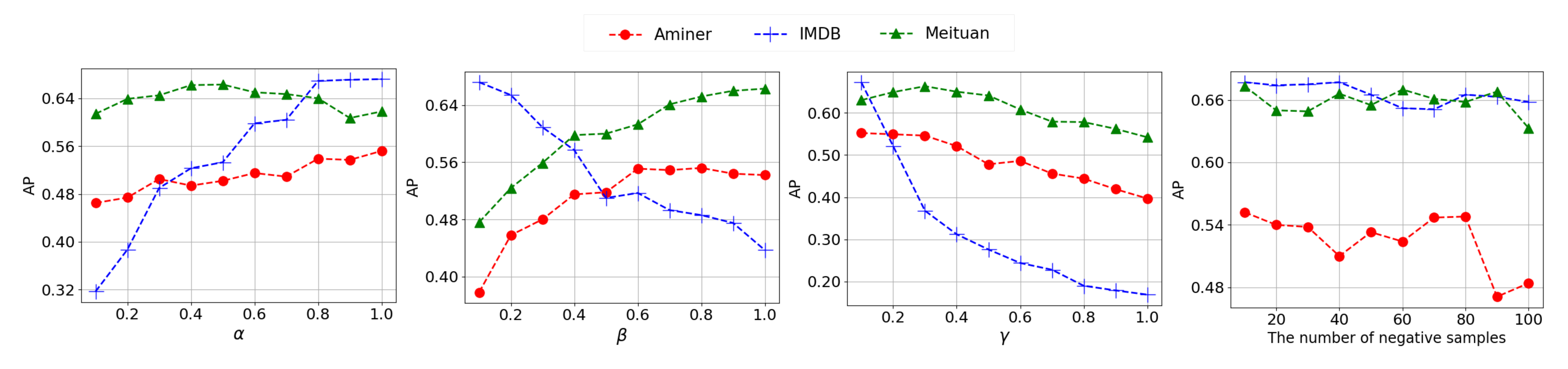}
    \caption{Parameters sensitivity analysis. (Left three figures: effects of different module weights $\alpha$, $\beta$ and $\gamma$; right figure: effects of different number of negative samples in pair-wise contrastive module.)}
    \label{fig:param}
\vspace{-5mm}
\end{figure*}

\subsection{Case Study}
We conduct a case study to show the detection results of AEHCL in Table \ref{tab:case_ab}. Note that we haven't reported the abnormal events in IMDB since we only have node features but no actual node information. In the case of the Aminer dataset, the three authors are all specialized in data mining. However, they collaborate to publish a text-processing paper about COVID-19 at a bio-medicine venue. We found that their regular pattern of this kind of collaboration is at least one biologist co-author. Therefore, though the pair-wise interactions in this event are normal, the multivariate interaction may be abnormal. In the case of Meituan dataset, the price of order A is greater than 50 and the food taste has a tag 1. However, the order is made by a user C whose historical average order price is less than 29 and the average taste tag is 2, which is an uncommon event. Also, user C usually orders food with category 3, but the shop in this order has category 1. In short, this is a common abnormal pair-wise interaction event.

\subsection{Model Analysis}
\label{sec:model Analysis}
In this section, we investigate the effect of different abnormal event score functions, as well as the performance of AEHCL under different hyper-parameters.

\stitle{Effect of Different Abnormal Score Functions.} 
The design of abnormal event score function can be divided into two parts. One is for the pair-wise contrastive module, and another is for the multivariate and inter-event contrastive module since these two parts use different forms of contrastive loss. For the pair-wise contrastive module, besides the $\textit{min}$ operation of our model, there are three other choices: opposite of average of node pair similarities (\textit{avg}), the standard deviation of node pair similarities (\textit{std}), and maximum of each node losses within the event using Equation (\ref{equ:pcm}) (\textit{loss}). We keep the other two modules unchanged and report the best results in Table \ref{tab:abnormal_score_pw}. The results show that the $\text{AEHCL}_{min}$ is not the best for all datasets (it performs a little worse on the Meituan dataset) but the most robust choice. Compared with others, the $\text{AEHCL}_{std}$ performs unstable on different datasets, which demonstrates this score setting is less discriminative in some scenarios.

As for the multivariate and inter-event contrastive module, similar to \cite{DBLP:journals/tnn/LiuLPGZK22}, we compare three abnormal event score methods: $\text{AEHCL}_+$ means that only the scores of positive samples are used to calculate abnormal event score, and $\text{AEHCL}_-$ denotes that only using negative samples. $\text{AEHCL}_{+/-}$ combines both positive and negative sample scores. We keep the pair-wise contrastive score unchanged and report the best performance in Table \ref{tab:abnormal_score_mul_inter}. Compared with $\text{AEHCL}_+$, the performance of $\text{AEHCL}_-$ decreases dramatically, indicates the negative sample scores have little difference between normal and abnormal events. This makes sense since the negative samples are all sampled in the same way. Combining two scores also harms the overall performance. We suppose that adding the negative contrastive score may disturb the true degree of event abnormality.

\begin{table}[!t]
\small
    \centering
    \fontsize{6.5pt}{8.5pt}\selectfont
    \renewcommand\arraystretch{1}
    \caption{Performance (\%) on different abnormal score functions in pair-wise contrastive module.}
    \resizebox{\columnwidth}{1.35cm}
	{
    \begin{tabular}{c|cc|cc|cc}
\hline
                          \textbf{Datasets}& \multicolumn{2}{c|}{\textbf{Aminer}}                                          & \multicolumn{2}{c|}{\textbf{IMDB}}                                            & \multicolumn{2}{c}{\textbf{Meituan}}                                         \\ \hline \hline
\textbf{Metrics}          & \multicolumn{1}{c|}{AP} & \multicolumn{1}{c|}{AUC} & \multicolumn{1}{c|}{AP} & \multicolumn{1}{c|}{AUC} & \multicolumn{1}{c|}{AP} & \multicolumn{1}{c}{AUC} \\ \hline

$\text{AEHCL}_{avg}$    & \multicolumn{1}{c|}{49.3}        & 87.6                             & \multicolumn{1}{c|}{66.5}      & 94.0                            & \multicolumn{1}{c|}{\textbf{67.9}}        & \textbf{91.9}                            \\ \hline
$\text{AEHCL}_{std}$ & \multicolumn{1}{c|}{52.1}       & 87.4                              & \multicolumn{1}{c|}{39.9}       & 86.3                              & \multicolumn{1}{c|}{67.5}      & 91.3                             \\ \hline
$\text{AEHCL}_{loss}$                & \multicolumn{1}{c|}{50.6}       & 85.3                             & \multicolumn{1}{c|}{59.9}      & 93.2                            & \multicolumn{1}{c|}{67.0}      & 90.5                            \\ \hline
$\text{AEHCL}_{min}$                & \multicolumn{1}{c|}{\textbf{55.2}}      & \textbf{88.6}                            & \multicolumn{1}{c|}{\textbf{67.2}}      & \textbf{95.1}                            & \multicolumn{1}{c|}{66.3}      & 90.8                            \\ 
\hline

\end{tabular}}
    \label{tab:abnormal_score_pw}
\vspace{-6mm}
\end{table}

\begin{table}[!t]
    \centering
    \fontsize{6.5pt}{8.5pt}\selectfont
    \renewcommand\arraystretch{1}
    \caption{Performance (\%) on different abnormal score functions in multivariate and inter-event contrastive module.}
    \resizebox{\columnwidth}{1.15cm}
	{
    \begin{tabular}{c|cc|cc|cc}
\hline 
                         \textbf{Datasets} & \multicolumn{2}{c|}{\textbf{Aminer}}                                          & \multicolumn{2}{c|}{\textbf{IMDB}}                                            & \multicolumn{2}{c}{\textbf{Meituan}}                                         \\ \hline \hline
\textbf{Metrics}          & \multicolumn{1}{c}{AP} & \multicolumn{1}{c|}{AUC} & \multicolumn{1}{c}{AP} & \multicolumn{1}{c|}{AUC} & \multicolumn{1}{c}{AP} & \multicolumn{1}{c}{AUC} \\ \hline
$\text{AEHCL}_+$                & \multicolumn{1}{c}{\textbf{55.2}}      & \textbf{88.6}                            & \multicolumn{1}{c}{\textbf{67.2}}      & \textbf{95.1}                   & \multicolumn{1}{c}{\textbf{66.3}}      & \textbf{90.8}                            \\ \hline
$\text{AEHCL}_-$    & \multicolumn{1}{c}{21.8}        & 76.0                            & \multicolumn{1}{c}{42.2}      & 93.3                            & \multicolumn{1}{c}{10.9}        & 68.5                            \\ \hline
$\text{AEHCL}_{+/-}$ & \multicolumn{1}{c}{41.3}       & 85.0                             & \multicolumn{1}{c}{65.7}       & 94.8                              & \multicolumn{1}{c}{60.0}      & 89.8                             \\ \hline

\end{tabular}}
\vspace{-6mm}
    \label{tab:abnormal_score_mul_inter}
\end{table}

\stitle{Parameters Sensitivity.}
We investigate the effects of the three key hyper-parameters
($\alpha$, $\beta$ and $\gamma$) in AEHCL. For each parameter, we change it from 0 to 1 with step 0.1 while fixing the other two parameters as the best ones. AP score is reported in Figure \ref{fig:param} (AUC score has a similar trend). There have following observations: (1) The pair-wise contrastive module is significant for abnormal event detection. The model performance consistently increases along with the increment of $\alpha$ on two datasets, and also achieves best when $\alpha$ is near 0.5 on Meituan dataset. This indicates that the pair-wise interaction anomalies may be more general and can be well captured by the pair-wise contrastive module. (2) The multivariate contrastive module has different effects on different datasets. Though an increasing trend of effects can be found when increasing $\beta$ on Meituan dataset, the effects suffer decreasing on the other two datasets when $\beta$ is too large, notably on IMDB dataset. The reason may be that these two datasets has fewer multivariate interaction anomalies. (3) The inter-event contrastive module plays an complementary role in detecting abnormal events. The performance has a decreasing trend on all datasets when increasing $\gamma$. However, as discussed in Section \ref{sec:overall}, the model performs better when $\gamma$ is relatively small (e.g., $\gamma=0.1$) than setting $\gamma=0$. Therefore, despite the relatively few inter-event anomalies, this module is indispensable for further improving the detection performance.


We also analyze the significance of negative sample numbers $n$ in the pair-wise contrastive module. Figure \ref{fig:param} shows AP score under different number of negative samples. The performance on IMDB and Meituan datasets are relatively stable with different $n$. however, the model suffers performance decay when $n$ is large ($n>80$) on the Aminer dataset. Since we have few limitations on the sampling strategy of negative samples, there may involve some entities whose features are similar to the target entities when $n$ is too large. Therefore, we set $n=10$ for all datasets. More sampling strategies are expected to study in further work.



\section{Conclusion}
In this paper, we explore the challenge problem of abnormal event detection in AHIN. Different from previous works focusing on detecting abnormal events with simple pair-wise patterns, we model events as hyperedges and proposed a hypergraph contrastive learning method, named AEHCL, which includes three contrastive modules to capture pair-wise, multivariate, and inter-event abnormal patterns. Moreover, an abnormal event score function is proposed to measure the abnormal degrees.  Extensive experiments on three real-world datasets demonstrate the effectiveness of AEHCL.

\vspace{-2mm}
\section{Acknowledgement}
This work is supported in part by the National Natural Science Foundation of China (No. U20B2045, 62192784, 62172052, 62002029, 62172052, U1936014) and BUPT Excellent Ph.D. Students Foundation (No. CX2021118). 

\bibliographystyle{mybib}
\bibliography{mybib}

\vspace{-2mm}

\end{document}